# Implementation of a FPGA-Based Feature Detection and Networking System for Real-time Traffic Monitoring

Jieshi Chen, Benjamin Carrion Schafer, Ivan Wang-Hei Ho
Department of Electronic and Information Engineering, The Hong Kong Polytechnic University, Hong Kong
13111714G@connect.polyu.hk, {b.carrionschafer, ivanwh.ho}@polyu.edu.hk

*Abstract*—With the growing demand of real-time traffic monitoring nowadays, software-based image processing can hardly meet the real-time data processing requirement due to the serial data processing nature. In this paper, the implementation of a hardware-based feature detection and networking system prototype for real-time traffic monitoring as well as data transmission is presented. The hardware architecture of the proposed system is mainly composed of three parts: data collection, feature detection, and data transmission. Overall, the presented prototype can tolerate a high data rate of about 60 frames per second. By integrating the feature detection and data transmission functions, the presented system can be further developed for various VANET application scenarios to improve road safety and traffic efficiency. For example, detection of vehicles that violate traffic rules, parking enforcement, etc.

## I. INTRODUCTION

The increase in population density in cities across the world has led to the rapid growth of vehicles and the development of wide public transportation systems. In such complicated transport networks, there has been a growing demand of real-time traffic monitoring and surveillance so that the road users can better understand current traffic condition, detect important information from the road as well as transmit and share the collected data for record and cooperative control.

In this paper, we discuss the implementation of a feature detection and networking system on FPGA board to achieve real-time data collection, pre-processing and delivery. Specifically, the developed system prototype uses a camera to capture the raw data from the traffic scene. It then detects and locates the position of cars' number plates in the scene and then transmits the collected data through the Ethernet port to the wireless networking device for data delivery and sharing in the vehicular ad-hoc network (VANET). The functions of on-road feature detection and networking can be applied to many application scenarios. For instance, to identify vehicles that violate traffic rules and conduct parking enforcement.

Usually, there are two ways to implement feature detection on FPGA. One is to combine both the hardware system and software application to achieve the detection and data processing functions; another way is to use solely the hardware system to conduct all the detection, tracking and data processing. For example, the authors in [1] implemented a Harris feature detector to detect and track the target object in real time, in which the processing of the frame-to-frame movement is accomplished through software. On the other hand, reference [2] used only the FPGA board to construct the whole system and developed an architecture of FPGA/DSP-based reprogrammable processor. In this paper, the hardware-based approach is exploited to assure that the feature detection can be conducted in a smooth and efficient way to accommodate the real-time requirement.

For the above prior works, the researchers only implemented the feature detection function without considering data transmission. In this paper, we present a system prototype that integrates both the feature detection and data transmission functions to make it an integrated solution for mobile surveillance from data collection, processing to delivery. In addition, another highlight of this prototype is that it uses high-level synthesis for designing the feature detection hardware module. The high-level synthesis designing method uses high-level language such as ANSI C to describe the expected behavior in the original codes and then automatically generate the relevant digital hardware that can implement such a defined behavior [3]. The algorithm is first described and tested in ANSI C. After verification, it is automatically translated to RTL codes which can be directly implemented in hardware projects [4]. Through this method, the design period can be largely shortened and it provides a friendly environment for the designer to modify, test and debug the system [5].

## II. SYSTEM ARCHITECTURE

### A. Brief structure of the system

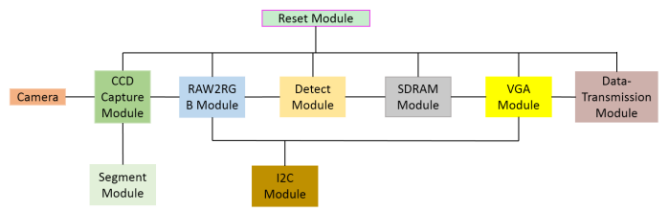

Figure 1.  Brief structure of the system.

The hardware platform for this design is the Altera DE2-115 development board. The camera used is TRDB-D5M. The software for constructing the design is Quartus II 13.0, and the tool for high-level synthesis is CyberWorkBench [5]. The hardware connections between different components are illustrated in Fig. 2.

The key modules of the whole system are illustrated in Fig. 1. The whole system undergoes several stages including optical signal capturing through the camera sensor; raw data collection through the CCD Capture module; data format transform through the RAW2RGB module; feature detection through the detect module; data storage and re-generation through the SDRAM module; visible output display through the VGA module; and Ethernet data transmission through the data transmission module. Other auxiliary modules like the I2C module is for adjusting the zoom in/out and light exposure of the camera, the segment module is for displaying the counted number of frames, and the reset module is for resetting the whole system.

The whole project is constructed using Quartus II 13.0 and Verilog codes to describe the behaviors of each module as well as the connections and relationships among them.

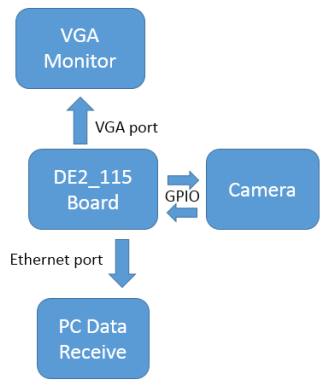

Figure 2.  Hardware connections between different components.

*B. Data format reconstruction*

The camera is inserted into the GPIO interface on the FPGA board. Pixel value of the raw data is 12 bits long. The data format from the output sensor side is in Bayer Pattern as shown in Fig. 3. Each raw data pixel from the sensor chip has only one color: red, green or blue. In order to represent a color pixel, the missing value is properly interpolated.

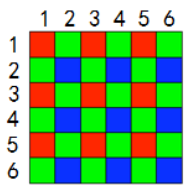

Figure 3.  Bayer pattern format of the raw pixel.

The main idea is to convert the former buffered Bayer color patterns to a 12-bit RGB value in real-time and conduct proper down-sampling so that the converted data can be processed in a later stage. To achieve this, the "altshift_taps" megafunction, which is commonly used in image processing program as data register, is used to create buffers for two taps of pixel data. The length of each tap is 1280, which equals to the length of each row in the raw data. For each pixel, it forms a 4-pixel template with its right, bottom, and bottom-right pixels for interpolation as illustrated in Fig. 4. The interpolation method is listed below.

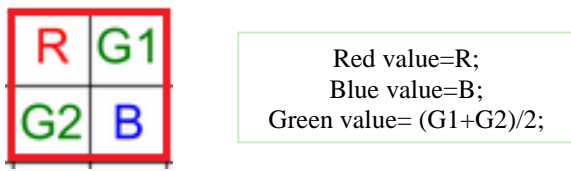

Figure 4.  RAW2RGB conversion equation.

## III. HIGH-LEVEL SYNTHESIS FOR FEATURE DETCTION

*A. Number-plate detection and location algorithm*

According to the characteristic and considering the efficiency of the hardware, the feature detection algorithm is designed to be pixel-based. Specifically, the algorithm scans the scene pixel-by-pixel and then line-by-line to look for the targeted feature similar to normal screen-scanning process. Therefore, one pixel is being processed by the detection module each time and judged whether it is the target pixel.

For instance, the background color of the number plate in Hong Kong is yellow, each yellow pixel will be considered as the potential target pixel for detection. The key idea of the algorithm is to find the first line of the number-plate and eliminate the noise pixels so as to accurately locate the plate.

To improve the detection accuracy of the yellow pixels, all the RGB pixel values are transformed into the HSV domain for processing. Each time when a yellow pixel is first found, its coordinates will be recorded and a counter will start to count the number of continuous yellow pixels and compare this value with a threshold. If the number exceeds the threshold, this line will be considered as the upper-line of the number plate, otherwise, it will be considered as noises. If the upper-line is found, one flag will be set and the algorithm will start to find the bottom line of the number-plate in a similar way. Hence, the area of the number plate can be effectively located, and a red rectangular block will be displayed to highlight the number plate area in the video output, while other parts in the scene will be eliminated as shown in Fig. 5.

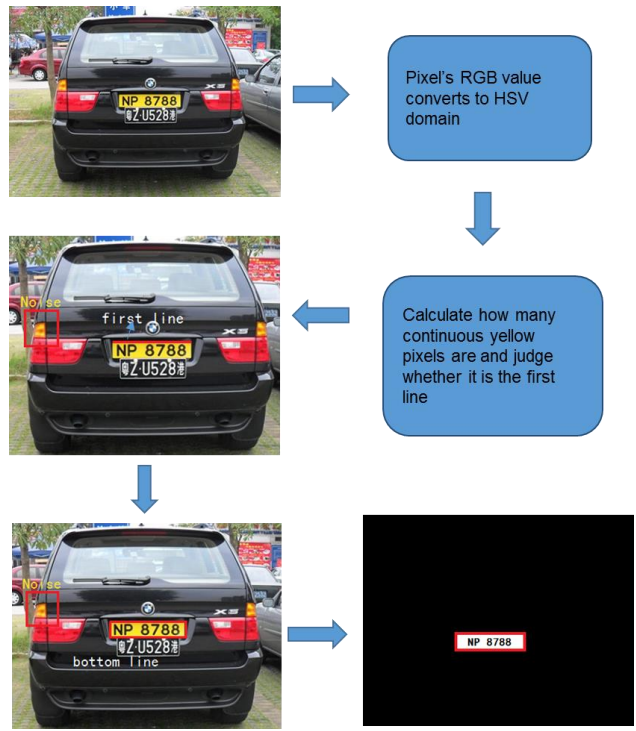

Figure 5.  Illustration of the number plate detection algorithm.

After the detection process, the whole picture will be represented in two colors only, white for the number plate background, and black for all the rest. Therefore, it largely reduces the amount of data which we need to transform and process in the system.

### B. High-level synthesis design and implemetation flow

The behavior of the detection algorithm was first written in ANSIC C in the CyberWorkBench (CWB) platform [5]. After successfully parsing the source C codes, several settings need to be made as to constrain the synthesis environment. To handle the high data rate in this design, the "automatic scheduling mode" is used as it will split the processing period into different stages and process the data in a parallel manner. Then, the source C codes will be high-level synthesized. The designer can review the quality and results of the synthesis by checking the synthesis reports, error reports, signal table, data path, etc.

Three verification steps then need to be done to verify if the algorithm performs correctly. The first step is *software simulation*. A bmp file is input to the simulator, which generates three files that record the pixel values of the test picture. These three files will be used as inputs to the next two verification steps. Furthermore, the software simulation also generates the result of the designed algorithm. This result will later be compared to the results of the other two verification steps. The second step is to do *cycle-accurate verification*. This step verifies the correctness of the timing and also the accuracy of the synthesized design. By looking at the wave stream of the simulation (Fig. 6), it is easier for the designer to identify any potential problems. The third step is to do *RTL-verification*. In this stage, the synthesizer simulates the true hardware environment and tests if the algorithm can run successfully. If the results of cycle-verification and RTL-verification match the result of software simulation, then we can assume that the algorithm is correct and the synthesized codes can perform the desired functions.

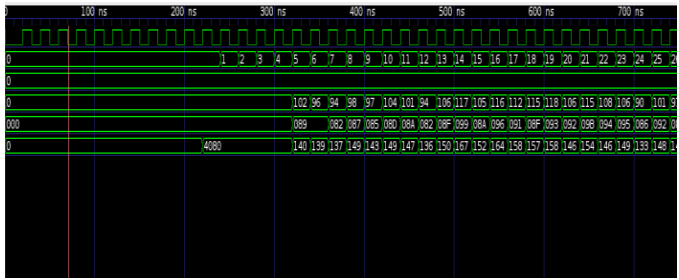

Figure 6. Waveform of the cycle-accurate simulation

On the other hand, if there is any mismatch discovered or mistake found, it is easy for the designer to modify the source C codes to revise the results. Moreover, according to the synthesis reports, the designer can carefully modify some parts of the source codes so as to improve the overall quality of the design, e.g., saving the area of the resource, or shortening the critical path time.

Thus, through high-level synthesis, it overrides the traditional RTL designing method for hardware projects and provides a more convenient way for the designer to test and modify the behavior of the system. The overall design flow of high-level synthesis is depicted in Fig. 7.

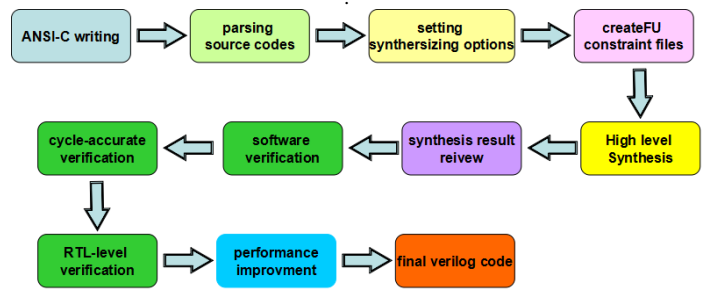

Figure 7. The design process on the CWB platform.

After the high-level synthesis process, the automatically generated Verilog codes can be directly implemented into the hardware project in Quartus II. Fig. 8 shows the operation of the number plate detection function that the system captures the image of the car and outputs the detected number plate on the monitor after image processing. It is evaluated that the system can process video at a rate up to 60 frames per second.

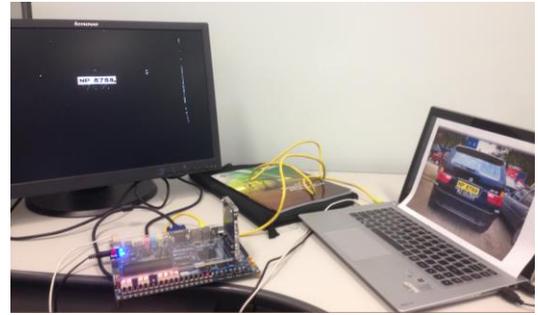

Figure 8. The operation of the number plate detection part.

## IV. DATA TRANSMISSION AND NETWORKING

### A. Data transmission design flow

After discussing the data collection and processing modules, how to send the processed data out to the wireless networking device so that detected data can be shared among other wireless users in the vehicular network is another major task of this design.

On the DE2-115 board, there is an "88E1111" chip for Ethernet data transmission. In this system, we choose to use the RGMII data transportation method to implement the 1000BASE-T Ethernet communication. Therefore, the standard transmission rate is 125MB/s. Furthermore, the UDP protocol is used for the delivery of the detected real-time data.

Since the data to be transmitted has only two colors, white (for the number plate background) and black (for the rest), it is reasonable to reconstruct the data format from 12 bits to 8 bits, which further reduces the overhead for Ethernet data transmission. Two FIFO queues are used for alternative read and write commands. After reconstructing the data, they will be sent to the FIFO queues and then encapsulated into data packages according to the UDP, IP and MAC protocols. A transmission module which executes the Ethernet data transmission will physically send out the data to other wireless networking devices (e.g., the IEEE 802.11p wireless modules for vehicle-to-vehicle communications). The design flow of the data transmission function is illustrated in Fig. 9.

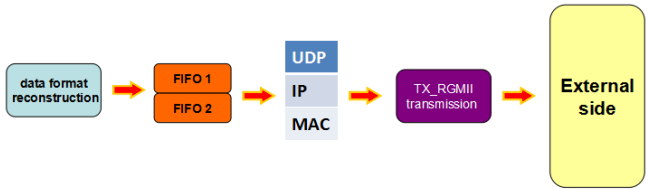

Figure 9. Ethernet data transmission design flow.

### B. Ethernet package structure

According to the standard UDP, IP, MAC and Ethernet packet structure, data frame which is running on the DE2-115 board can be defined. Table I shows the user defined Ethernet frame which is used in this system to transmit the real-time data through the Ethernet port.

TABLE I. USER DEFINED ETHERNET FRAME STRUCTURE.

| Byte number | Byte length | Value | definition | comments |
|---|---|---|---|---|
| 0-6 | 7 | h'55 | preamble | Frame header 8 bytes |
| 7 | 1 | h'D5 | delimiter | |
| 8-13 | 6 | | Destination MAC address | Ethernet II type 14 bytes |
| 14-19 | 6 | | Source MAC address | |
| 20 | 1 | h'08 | Type IP=h'8808 | |
| 21 | 1 | h'88 | | |
| 22 | 1 | h'45 | version/IP header length=20 bytes | IP header 20 bytes |
| 23 | 1 | h'00 | unknown TOS | |
| 24-25 | 2 | h'041c | total length of IP header and data=20+8+1024=0x41c | |
| 26-27 | 2 | h'0000 | adds one when on package is sent out | |
| 28-29 | 2 | h'0000 | FLAGS/offset | |
| 30 | 1 | h'80 | TTL | |
| 31 | 1 | h'11 | UDP protocol | |
| 32-33 | 2 | | IP header checksum | |
| 34-37 | 4 | | IP source address | |
| 38-41 | 4 | | IP destination address | |
| 42-43 | 2 | | port source address | UDP header 8 bytes |
| 44-45 | 2 | | port destination address | |
| 46-47 | 2 | h'408 | total length of UDP header and data=8+1024=1032 0x408 | |
| 48-49 | 2 | | UDP header checksum | |
| 50-1073 | 1024 | | data | 1024 byte |
| 1074-1077 | | | CRC | 4 byte |

### C. Ethernet data transmission module design

#### 1) Data format reconstruction module

The transmitted data is reconstructed before it enters the final transmission session. As mentioned above, all the output pixels have only two colors, black or white. This largely reduces the data amount which is needed to be transmitted and largely simplifies the data transmission procedure. From the VGA side, the output data is 10 bits long. Since the Ethernet usually sends out data based on bytes (8 bits per byte), we reconstructed the data format so that it is easier to be encapsulated in the data packet.

After the data reconstruction process, data are sent to the FIFO queues. One of the queues is for writing and the other is for reading. These two queues work alternatively to support the whole system's read/write requirements.

#### 2) Data generation module

This module is the key component of the data transmission part. Within this module, all the MAC layer, IP layer and UDP layer headers are constructed. It generates data at a rate of 125MB/s, which is the standard gigabit Ethernet transmission rate. The data frames are encapsulated according to the Ethernet frame II data package definition. Following this procedure, CRC checksum is also conducted.

#### 3) Data transmission module

After getting through all the three layers and passed the CRC checksum, the data packet is ready for transmission. The final sending module, which is directly connected to the Ethernet PHY chip, is composed of three parts realized through mega functions.

*Sending signal control*: This module is for generating the clock pulse. Data sending is based on this clock signal.

*Data sending*: This module is for operating the external Ethernet PHY chip to transmit data.

*Double-speed sending realization*: This module is for realizing double-speed data transmission to improve the sending speed. Usually, data are only sent at the positive edge of the clock, but in this double-speed sending module, data are sent at both the positive and negative edges of the clock as illustrated in Fig. 10. Through this module, at the positive edge of the clock pulse, 4 high-order bits will be sent; at the negative edge of the clock pulse, 4 low-order bits will be sent.

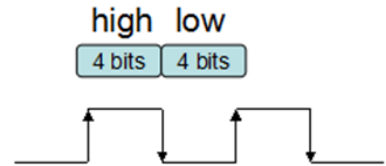

Figure 10. Illustration of double-speed data transmission.

### D. Data reception

An 802.11 wireless device is connected to the FPGA board through the Ethernet interface for data reception. To successfully receive data, specific MAC address and IP address should be pre-defined. On the FPGA side, the target MAC address should be the connected device's MAC address. On the connected device, the IP address should be set to a pre-defined domain, for example, 192.168.1.0. Data received by the wireless device can be broadcasted to other wireless users in the vicinity for information dissemination in the vehicular network and storage at central depository for future analysis.

## V. DISCUSSION AND CONCLUSION

### A. Resource consumption of different modules

Fig. 11 and 12 illustrate the resource occupation of each module and the area consumption of the whole design on the FPGA board, including the number of logic elements used on the board, the I/O pin usage, the memory bits consumption, etc. The detection module, as being the most complicated module within the system, consumes most resources comparing to other modules (Fig. 11). It is then followed by the memory module and the Ethernet transmission module. Overall, the proposed system consumes quite few resources on the FPGA board to achieve its desired functions (only 4% of the total elements on the board) as shown in Fig 12.

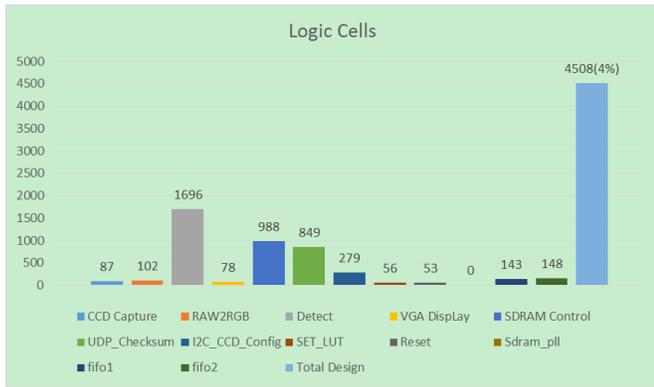

Figure 11. Logic cells resource occupation.

Figure 12. Resource usage summary report.

### B. Conclusion

In this paper, we discuss the implementation of a hardware-based feature detection and data transmission system prototype on FPGA for real-time traffic monitoring. The system exploits high-level synthesis to reduce the development effort and enable more flexible modification through high-level languages like ANSI C. We proposed and implemented a number plate detection algorithm on FPGA that can tolerate a high data rate of about 60 frames per second. In addition, we also enabled the data transmission function on the FPGA so that the detected information can be delivered to other locations in the communication network for storage and analysis. Overall, the integrated system can conduct data collection, image processing and data transmission in a real-time manner, which can be further developed and applied in various VANET application scenarios such as parking enforcement and traffic violation detection.

ACKNOWLEDGMENT

This work is partially supported by the Department of Electronic and Information Engineering, The Hong Kong Polytechnic University (Projects 4-ZZDF and 4-ZZCZ).